\documentclass[conference]{IEEEtran}

\usepackage{cite}
\usepackage{amsmath,amssymb,amsfonts}
\usepackage{algorithmic}
\usepackage{graphicx}
\usepackage{textcomp}
\usepackage{xcolor}
\usepackage{hyperref}

\def\BibTeX{{\rm B\kern-.05em{\sc i\kern-.025em b}\kern-.08emT\kern-.1667em\lower.7ex\hbox{E}\kern-.125emX}}

\definecolor{tabR}{HTML}{9C0006}
\definecolor{tabG}{HTML}{006100}
\definecolor{tabY}{HTML}{FFD700}

\begin{document}

\title{Anticipate, Ensemble and Prune: \\Improving Convolutional Neural Networks via Aggregated Early Exits}


\author{
    
    \IEEEauthorblockN{Simone Sarti}
        \IEEEauthorblockA{
        \textit{DEIB, Politecnico di Milano}\\
        Milan, Italy \\
        simone.sarti@mail.polimi.it
    }
    
    \and
    
    \IEEEauthorblockN{Eugenio Lomurno}
    \IEEEauthorblockA{
        \textit{DEIB, Politecnico di Milano}\\
        Milan, Italy \\
        eugenio.lomurno@polimi.it
    }
    
    \and
    
    \IEEEauthorblockN{Matteo Matteucci}
        \IEEEauthorblockA{
        \textit{DEIB, Politecnico di Milano}\\
        Milan, Italy \\
        matteo.matteucci@polimi.it
    }
}

\maketitle

\begin{abstract}

Today, artificial neural networks are the state of the art for solving a variety of complex tasks, especially in image classification. 
Such architectures consist of a sequence of stacked layers with the aim of extracting useful information and having it processed by a classifier to make accurate predictions.
However, intermediate information within such models is often left unused. In other cases, such as in edge computing contexts, these architectures are divided into multiple partitions that are made functional by including early exits, i.e. intermediate classifiers, with the goal of reducing the computational and temporal load without extremely compromising the accuracy of the classifications.
In this paper, we present Anticipate, Ensemble and Prune (AEP), a new training technique based on weighted ensembles of early exits, which aims at exploiting the information in the structure of networks to maximise their performance. Through a comprehensive set of experiments, we show how the use of this approach can yield average accuracy improvements of up to 15\% over traditional training. In its hybrid-weighted configuration, AEP's internal pruning operation also allows reducing the number of parameters by up to 41\%, lowering the number of multiplications and additions by 18\% and the latency time to make inference by 16\%. By using AEP, it is also possible to learn weights that allow early exits to achieve better accuracy values than those obtained from single-output reference models.
\end{abstract}


\section{Introduction}\label{sec:introduction}

Over the last decade, deep learning has emerged as one of the dominant disciplines in computer science, thanks to the research conducted and the remarkable results obtained.
Among the main fields of application, that of visual recognition, and in particular that of image classification, has undoubtedly been the catalyst for a veritable revolution, rooted in the development and refinement of architectures known as convolutional neural networks (ConvNets).
Such artificial neural networks involve numerous convolutional layers that extract the information contained in input images and process it to obtain high-level features. 

The development of the first ConvNet, called AlexNet~\cite{krizhevsky2017imagenet}, was made possible by the increasing production and storage of data and, in particular, by the public release of the ImageNet benchmark~\cite{russakovsky2015imagenet}. From then on, the succession of discoveries of increasingly high-performance models accelerated.
Among the major milestones, architectures such as VGG~\cite{simonyan2014very}, Inception\cite{szegedy2015going}, ResNet~\cite{he2016deep}, DenseNet~\cite{huang2017densely}, MobileNet~\cite{howard2019searching}, EfficientNet~\cite{tan2019efficientnet} and recently ConvNeXt~\cite{liu2022convnet} excelled in setting new levels in terms of accuracy, scalability, efficiency and design quality.

Recently, the study of neural networks with early exits has gained importance. An early exit of a neural network is a classifier placed at an intermediate level between the input layer and the traditional single output layer. The objectives of such a design pattern are many and varied, including exploiting the information contained in the intermediate layers of the models, streamlining their overall weight by cutting them, or for purposes related to distributed systems and edge computing~\cite{matsubara2022split}.
The optimal number of branches with early exits and their positioning represent an important choice in this context, especially for very deep ConvNets or for architectures that are not strictly sequential.
Another fundamental step lies in the choice of updating the weights of such architectures with multiple outputs, their individual or joint use, and the management of outputs with degraded performance.

This work shows the analysis of the behaviour of the main ConvNets presented in the literature modified through the proposed early exits technique named Anticipate, Ensemble and Prune (AEP).
In particular, AEP is presented as an early exits-weighted ensemble technique. Outputs aggregation strategies for both loss function and inference are discussed, in order to understand which conditions favour an improvement or lead to a loss in classification performance compared to the basic single-output version of the neural networks examined.
Finally, through the adoption of a pruning step, it is shown how it is possible to reduce the number of parameters, operations and network latency, further increasing accuracy through the extraction of the optimal sub-ensemble network.
The experiments are conducted on a large set of ConvNets and datasets and both in a traditional training context and through the tuning of pre-trained architectures.
Unlike the main reference works in the literature, which aim to reduce latency as much as possible without sacrificing model accuracy or to develop techniques for edge computing~\cite{wolczyk2021zero, sun2021early, campbell2022robust}, this work aims to quantify and maximise the accuracy gain that the use of early exits' ensemble can provide. 

The document is further divided into five sections. 
The Section~\ref{sec:related} summarises related work that has been proposed in the literature and is useful for understanding the rest of the document. 
The Section~\ref{sec:proposed_approach} describes the steps that make up the AEP technique. 
The section~\ref{sec:setup} describes the details of the experiments and the configurations with which they were performed.
The section~\ref{sec:results} presents and discusses the results of the experiments. 
Finally, the section~\ref{sec:conclusion} concludes the work and suggests some possible research directions.

\section{Related Works}\label{sec:related}

Early exits is a deep learning technique that aims to improve the efficiency of neural network models by allowing them to make predictions before the input data is processed by all the available layers. This is done by training multiple sub-models within the backbone model, each with a different level of complexity.
One of the first works in which the early exits technique is used was carried out by Panda \textit{et al.} and applied to convolutional neural networks (ConvNets)~\cite{panda2016conditional}. In particular, the aim of this study was to create an algorithm capable of identifying the optimal depth within the classification network under examination, so that the computational expense could be dynamically adjusted without losing accuracy.
During the same period, Teerapittayanon \textit{et al.} presented their work on early exits aimed at demonstrating the predictive properties of classifiers placed in the intermediate layers of ConvNets, such that the easiest samples to predict are processed by fewer hidden layers, while the most difficult ones traverse the entire architecture~\cite{teerapittayanon2016branchynet}.

A more recent approach aimed at reducing the energy cost and complexity of single-output ConvNets has been proposed by Wang \textit{et al.} achieving an extremely beneficial trade-off between accuracy and flops~\cite{wang2019dynexit}. The technique implemented involves the use of early exits and weighted loss functions applied to architectures containing skip connections.
Pacheco \textit{et al.} demonstrated how the use of early exit architectures is incredibly beneficial in the context of edge computing~\cite{pacheco2021towards}. In particular, they shown how the ability to classify non-anomalous samples at the shallow levels of a ConvNet allows not losing performance compared to a single-exit classification.

Having several available classifiers, some of them not necessarily high-performing, the most intuitive step is to identify an intelligent aggregation strategy to exploit their joint potential.
Ensembling, largely seen as the natural solution in many classification problems, is a machine learning technique that consists of training several different models to solve the same task and then exploiting the knowledge derived from all of them at inference time to make the best choice. The ensemble technique works because the different models have different weaknesses that are compensated by the others' strength points. 
Ensembling of early exits consists in classifiers sharing part of their structure and parameters, but still working on different features given the same input data.
This technique has been exploited by Wo{\l}czyk \textit{et al.} to produce an early exit-based approach in which each prediction is reused by subsequent exits, combining previous results in an ensemble-like manner~\cite{wolczyk2021zero}. The goal of this work was to minimise the prediction latency without sacrificing the accuracy performance of the proposed models.

For what concerns the training of early exit networks, joint training is the most common approach and consists in formulating a single optimization problem whose loss depends on all branches, in particular the network loss is often calculated as the weighted sum of the branches losses ~\cite{scardapane2020should,teerapittayanon2016branchynet,qendro2022towards}. Strategies have also been devised to dynamically adjust the exits losses weights during training~\cite{wang2019dynexit} and to improve the ensemble by combining the prediction loss with a a diversity loss~\cite{qendro2021early, sun2021early}. When early exits are trained jointly with the backbone network, they favor the learning of more discriminative features at each layer and lead to faster convergence, while also acting as regularization~\cite{lee2015deeply, teerapittayanon2016branchynet}.

The output of early exits ensemble can also be computed in a multitude of ways, e.g., as arithmetic mean of the predictions~\cite{qendro2021early, qendro2022towards}, via geometric ensemble~\cite{wolczyk2021zero} or through voting strategies~\cite{sun2021early}. 
The effectiveness of early exits ensembles is not limited to image classification, as they've also recently been used for image captioning~\cite{fei2022deecap}, natural language processing~\cite{sun2021early}, for uncertainty quantification and biosignal classification~\cite{qendro2021early, campbell2022robust} and to improve robustness against adversarial attacks~\cite{qendro2022towards}. Moreover, early exits ensembles were employed to produce a teacher-free knowledge distillation technique by treating the aggregated predictions as the teacher predictions~\cite{lee2021students}.

\section{Method}\label{sec:proposed_approach}

\begin{figure}[t]
\centering
\includegraphics[scale=0.16]{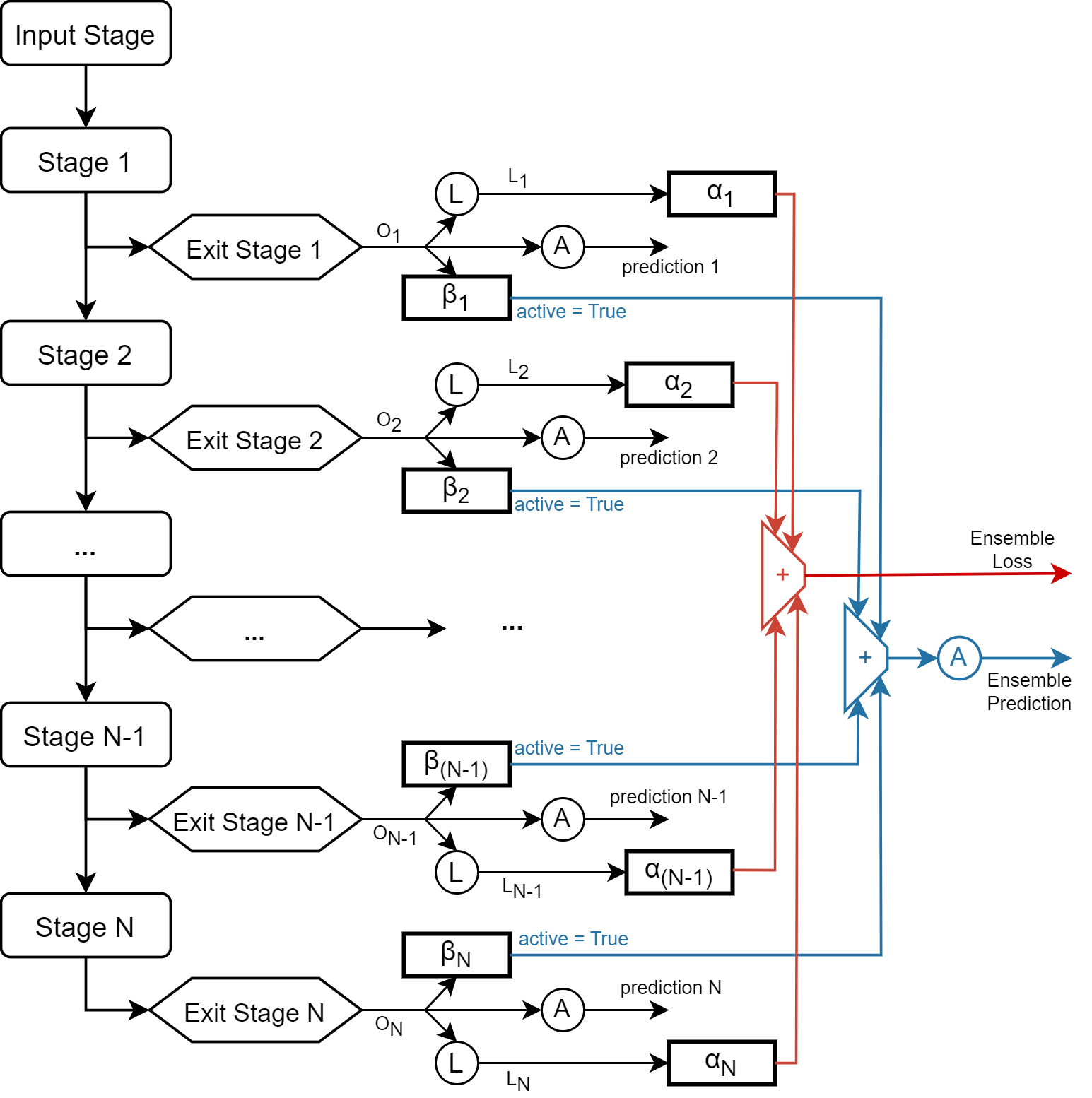}
\caption{Outline of the AEP technique. $O_i$ represents the output of the  i-th exit stage with linear activation. \raisebox{.5pt}{\textcircled{\raisebox{-.9pt} {$L$}}} epresents the loss function, i.e., the categorical cross-entropy, while \raisebox{.5pt}{\textcircled{\raisebox{-.9pt} {$A$}}} represents the argmax function used to obtain the class prediction. $\alpha_i$ and $\beta_i$ represent the weights assigned to the loss and output of exit stage $i$, respectively. The activation of the outputs is relative to the pruning step with which the proposed method terminates.}
\label{fig:method_schema}
\end{figure}

The technique of ensemble based on early exits proposed in this work, and called Anticipate, Ensemble and Prune (AEP), has been fully tested in the computer vision field to solve image classification tasks, but can easily be extended to other types of data and purposes.
Given a ConvNet, it is practically always possible to identify the presence of repeating blocks or stages stacked within it. Regardless of the architecture in question, it is therefore possible to replicate the classifier present at the end of the ConvNet, i.e. after the last stage, immediately after each intermediate stage, as can be seen in Figure~\ref{fig:method_schema}.

The training of an AEP model proceeds as follows: for each batch of images with dimension $B$, given $C$ target classes, the algorithm lets the images flow through the network such that every classifier outputs a tensor of $B$ elements, each representing one image and containing $C$ values, one for each class $c$. In this setting, the network loss is computed as the weighted sum of the categorical cross entropy loss $L_i$ of each exit $i \in [1,N]$, in a joint training fashion, as in Equation~\ref{eq:cce}. 

\begin{equation}
    L_{i} = L_{cce,i} = -\frac{1}{B}\sum_{b=1}^B\sum_{c=1}^C(p_{b,c}\log(y_{b,c}))
\label{eq:cce}
\end{equation}

Unlike the other approaches, in AEP the last exit is not treated differently from the intermediate ones, since its contribution to the final loss is weighted following the same weight assignment strategy as the others. 
The Equation~\ref{eq:loss} represents the network loss, where $\alpha_i$ is the weight assigned to the loss associated with output $i$.

\begin{equation}
    L = \sum_{i=1}^{N} \alpha_i \cdot L_{i}
\label{eq:loss}
\end{equation}

The predictions obtained after the ensemble, from now on referred to as $\hat{y}$, are calculated as a weighted sum of the outputs of each i-th output, namely $O_i$. The calculation of each classification metric is performed after applying the argmax operator to the vector thus obtained.
The Equation~\ref{eq:output} represents the prediction step, where $\beta_i$ is the weight assigned to the outputs associated with exit $i$.

\begin{equation}
    \hat{y} = \operatorname*{argmax} (\sum_{i=1}^{N} \beta_i \cdot O_i)
\label{eq:output}
\end{equation}

The strategy for selecting weights $[\alpha_1, ... , \alpha_N]$ and $[\beta_1, ... , \beta_N]$ is a crucial step in AEP.
To make the selection independent of the choice of backbone neural network and dataset, the weights of the loss function as well as of the prediction ensemble were tested in linearly increasing or decreasing form. Furthermore, the values of these weights were chosen such that they were strictly positive and with sum equal to one. In this way, it is possible to maintain the desired properties by having adaptive scales of weights regardless of the number of early exits within the neural network.

In the initial set of experiments, the same weights were used to compute both the network loss and the output of the ensemble. Specifically, the weights were either always increasing or always decreasing. To differentiate between these two cases, the experiments in which the weights were assigned in a descending order for both the losses and the outputs were referred to as "EEdesc", while those in which the weights were assigned in an ascending order for both the losses and the outputs were referred to as "EEasc".
Ablation studies demonstrated that, while the use of decreasing weights generally led to better exits when considered individually, particularly for earlier exits, networks with ascending weights performed better in terms of ensemble prediction. To address this, the two sets of weights were decoupled, and "EEmix" networks were tested, in which the exits losses weights were assigned in descending order and the outputs weights were assigned in ascending order.
Additionally, a uniform weight mode was tested, in which all exits were given the same weight. These experiments were referred to as "EEunif". A summary of the different weight modes can be found in Table~\ref{table:weights_modes}.

\begin{table}[t]
\vspace{-4mm}
\caption{The list of weight configurations for loss functions and exits predictions.}
\begin{center}
\begin{tabular}{|ccc|}
\hline
\textbf{Weights Mode} & \textbf{Losses Weights} & \textbf{Outputs Weights} \\ \hline \hline
DESC                     & decreasing              & decreasing              \\ \hline
ASC                      & increasing              & increasing              \\ \hline
MIX                      & decreasing              & increasing              \\ \hline
UNIF                     & uniform                 & uniform                 \\ \hline
\end{tabular}
\label{table:weights_modes}
\end{center}
\vspace{-3mm}
\end{table}

Upon completion of training the multi-output network, a pruning step is applied as a post-processing technique to optimize its performance. The resulting ConvNet is loaded and validated in all possible combinations of exit activation states, using the same exit weights used in training. The best sub-network is then extracted and evaluated on the test set. This selection process can aid in further improving accuracy in comparison to both the full ensemble and the single-exit network, while also reducing the number of parameters, operations, and latency as entire sections of the original network can be removed.

\section{Experiments Setup}\label{sec:setup}

The experiments conducted to evaluate the effectiveness of the AEP technique included several ConvNet architectures and many of the major image classification benchmarks. For an analysis characterized by high completeness, all experiments were evaluated with two different input scales, and each ConvNet was trained both through training from scratch and through fine-tuning from the weights learned on ImageNet.

\subsection{Networks}

The AEP technique was tested on 5 well known network architectures, thus encompassing many of ConvNets most relevant architectural trends and patterns.
The goal was to span different network sizes, design patterns and inner components.
An abstract and general architecture was constructed as a theoretical model for the various ConvNets considered, observable in Figure~\ref{fig:method_schema}.
By identifying the recurring cells in these architectures, it was possible to identify the list of stages to which early exits should be attached. At this point, each network could be enriched with the function to extract a subnetwork by specifying which exits to preserve. Below the list of networks identified and used in the experiments:

\begin{itemize}
    \item \textbf{ResNet50}~\cite{he2016deep}: 
    In terms of size, it is in the middle of the ResNet family of architectures. ResNets are characterized by the use of skip/residual connections between layers, which should help in improving gradient flow through the network. 
    \item \textbf{VGG16}~\cite{simonyan2014very}:  
    It is a purely sequential ConvNet, and while smaller than other VGG models, it is still by far the most demanding in terms of operations required.
    \item \textbf{DenseNet169}~\cite{huang2017densely}: 
    It is found in the middle of the DenseNet family of architectures, they are characterized by a dense blocks structure, in which features obtained after a layers are not only passed to the successive layers but also concatenated to the outputs of said layers, so that the information in a block is better preserved and exploited.
    \item \textbf{MobileNetV3Small}~\cite{howard2019searching}: 
    Designed to work under mobile settings, it was a good candidate to see what would happen with tiny architectures. It is also characterized by the use of a particular type of block called the Residual Inverted Bottleneck, characterized by high memory efficiency.
    \item \textbf{EfficientNetB5}~\cite{tan2019efficientnet}: 
    Part of the EfficientNet family, currently one of the best performing architecture families, this model in particular was chosen because with its more than 500 layers and, in contrast with MobileNetV3Small, allowed to study a huge architecture.
\end{itemize}

For each of the 5 networks, the classifier present in the original single-exit architecture was extracted and replicated for each of the early exits. The only exit stage altered from the original structure was that of the VGG16 model, which by default is extremely large and over-parameterized. It was substituted it with a simpler Global Average Pooling layer followed by a dense layer with a number of units equal to the number of possible classes.

The experiments conducted considered each network both with a number of early exits set at 4, for comparison purposes, and with a number of early exits equal to the number of stages in the original model. Specifically, while the experiments on Resnet50 and DenseNet169 did not need to be repeated since 4 was already the correct number of exits to match the number of stages, VGG16 and MobileNetV3Small required also a set of experiments concerning 5 exits, while EfficientNetB5 required also tests involving 7 exits. We will refer to these 3 networks by adding the “full” keyword to their name: \textbf{VGG16full}, \textbf{MobileNetv3Smallfull} and \textbf{EfficientNetB5full}, bringing the total number of networks tested to 8.

\subsection{Datasets and Metrics}

In this study, a thorough experimental evaluation of the AEP algorithm was conducted using six diverse image classification datasets. These datasets were chosen to cover a range of characteristics, including variations in class imbalance and image modality, in order to assess the generalizability and robustness of the proposed algorithm. The datasets used in this study spanned a range of class sizes, including datasets with tens to hundreds of classes, and are summarized in Table~\ref{table:datasets_properties}.

\begin{table}[t]
\caption{The list of datasets used to conduct the experiments with their characteristics.}
\begin{center}
\begingroup
\setlength{\tabcolsep}{0.45\tabcolsep}
\begin{tabular}{|c||c|c|c|c|c|c|}
\hline
\textbf{Dataset}     & \textbf{Classes}  & \textbf{Balanced}        & \textbf{Channels}   & \textbf{Train}     & \textbf{Validation}       & \textbf{Test} \\ \hline \
Cifar10\cite{krizhevsky2009learning}    & 10  & yes          & RGB          & 45000            & 5000       & 10000     \\ \hline
Cifar100\cite{krizhevsky2009learning}   & 100 & yes        & RGB            & 45000            & 5000       & 10000     \\ \hline
Eurosat\cite{helber2019eurosat}         & 10  & no         & RGB            & 17500             & 4000       & 5500      \\ \hline
FMNIST\cite{xiao2017fashion}            & 10  & yes        & Grayscale      & 51000             & 9000       & 10000     \\ \hline
GTSRB\cite{stallkamp2012man}            & 43  & no         & RGB            & 33209             & 6000       & 12630     \\ \hline
TinyImg\cite{le2015tiny}                & 200 & yes        & RGB            & 85000             & 15000      & 10000     \\ \hline
\end{tabular}
\endgroup
\end{center}
\label{table:datasets_properties}
\end{table}

The evaluation included comparing the performance of baseline and ensemble networks under different scenarios, including training from scratch and fine-tuning from ImageNet weights for all networks as well as using images of sizes 224x224 and 64x64 for all datasets. For each experiment, a set of performance metrics were collected, including Top1 accuracy, number of parameters, number of MACs (or Multiply-ACcumulate operations), latency, and training time. Additionally, for multi-output networks, the Top1 accuracy for each exit was also recorded. 

\subsection{Hyperparameters Setting}

The goal of this research was to evaluate the performance of single-exit networks in comparison to early-exit ensembles rather than matching or beating state of the art results. To accomplish this, a simple training approach was employed, using 100 training epochs, a batch size of 64, and a learning rate of $10^{-4}$ with the Adam optimizer. The other training parameters were left at their default values in PyTorch. 
An early stopping technique was applied, with a patience of 12 epochs, based on the validation loss. 
No data augmentation or learning rate schedules were utilized. 
The only preprocessing applied to the images was resizing to 224 or 64 through bicubic interpolation and normalization. The small learning rate was chosen to ensure that the same hyperparameters could be used in fine-tuning, thereby ensuring comparable results. The models were implemented using PyTorch and executed on an NVIDIA Quadro RTX 6000 GPU.

\section{Results and Discussion}\label{sec:results}

\begin{table*}[t!]
\caption{Comparisons between models based on early exits and those based on single exits, calculated as average percentage change of accuracy. The four experiment configurations are sorted by row according to the training technique and the size of the input images. The results are grouped by architecture (left column) and by dataset (right column). Improvements are shown in green, worsenings in red. The best results are highlighted in bold.}
\vspace{5mm}

\hrule
\begin{center}
\textbf{TRAIN-224}
\end{center}
\hrule
\vspace{0.4cm}
\begin{minipage}{.5\linewidth}

\fontsize{6.5}{8}\selectfont
\begingroup
\setlength{\tabcolsep}{0.45\tabcolsep}
\begin{tabular}{|c||c|c|c|c||c|c|c|c|}
\hline
 \textbf{Network} & \textbf{EEdesc}     & \textbf{EEasc}& \textbf{EEmix}& \textbf{EEunif}     & \textbf{EEdesc*}    & \textbf{EEasc*}     & \textbf{EEmix*}     & \textbf{EEunif*}    \\ \hline \hline
\textbf{Resnet50}     &\textcolor{tabG}{7,962}    &\textcolor{tabG}{7,948}    &\textcolor{tabG}{9,901}    &\textcolor{tabG}{8,510}    &\textcolor{tabG}{8,617}    &\textcolor{tabG}{7,997}    &\textcolor{tabG}{\textbf{10,088}}   &\textcolor{tabG}{8,831}    \\ \hline
\textbf{Vgg16}  &\textcolor{tabG}{0,849}    &\textcolor{tabG}{\textbf{4,653}}    &\textcolor{tabG}{3,670}    &\textcolor{tabG}{4,360}    &\textcolor{tabG}{2,596}    &\textcolor{tabG}{4,651}    &\textcolor{tabG}{3,688}    &\textcolor{tabG}{4,407}    \\ \hline
\textbf{Vgg16full}    &\textcolor{tabG}{0,922}    &\textcolor{tabG}{4,063}    &\textcolor{tabG}{3,517}    &\textcolor{tabG}{4,035}    &\textcolor{tabG}{3,012}    &\textcolor{tabG}{4,217}    &\textcolor{tabG}{3,524}    &\textcolor{tabG}{\textbf{4,859}}    \\ \hline
\textbf{DenseNet169}  &\textcolor{tabR}{-1,335}   &\textcolor{tabG}{0,804}    &\textcolor{tabR}{-0,066}   &\textcolor{tabG}{0,395}    &\textcolor{tabR}{-0,410}   &\textcolor{tabG}{0,803}    &\textcolor{tabG}{0,046}    &\textcolor{tabG}{\textbf{1,046}}    \\ \hline
\textbf{MBV3-small}  &\textcolor{tabG}{12,046}   &\textcolor{tabG}{9,672}    &\textcolor{tabG}{11,644}   &\textcolor{tabG}{9,882}    &\textcolor{tabG}{\textbf{12,289}}   &\textcolor{tabG}{9,588}    &\textcolor{tabG}{11,679}   &\textcolor{tabG}{10,055}   \\ \hline
\textbf{MBV3-smallfull}   &\textcolor{tabG}{11,361}   &\textcolor{tabG}{9,803}    &\textcolor{tabG}{12,727}   &\textcolor{tabG}{10,249}   &\textcolor{tabG}{\textbf{13,661}}   &\textcolor{tabG}{9,817}    &\textcolor{tabG}{12,650}   &\textcolor{tabG}{10,591}   \\ \hline
\textbf{EffNetB5}     &\textcolor{tabG}{1,630}    &\textcolor{tabG}{2,100}    &\textcolor{tabG}{2,108}    &\textcolor{tabG}{\textbf{2,362}}    &\textcolor{tabG}{1,885}    &\textcolor{tabG}{2,113}    &\textcolor{tabG}{2,239}    &\textcolor{tabG}{2,086}    \\ \hline
\textbf{EffNetB5full} &\textcolor{tabG}{3,710}    &\textcolor{tabG}{3,046}    &\textcolor{tabG}{4,443}    &\textcolor{tabG}{4,320}    &\textcolor{tabG}{4,065}    &\textcolor{tabG}{2,958}    &\textcolor{tabG}{\textbf{4,480}}    &\textcolor{tabG}{4,359}    \\ \hline \hline
\textbf{Total}  &\textcolor{tabG}{4,643} &\textcolor{tabG}{5,261} &\textcolor{tabG}{5,993} &\textcolor{tabG}{5,514} &\textcolor{tabG}{5,714} &\textcolor{tabG}{5,268} &\textcolor{tabG}{\textbf{6,049}} &\textcolor{tabG}{5,779} \\ \hline
\end{tabular}
\endgroup
  
\end{minipage}%
\hspace{0.35cm}
\begin{minipage}{0.5\linewidth}

\fontsize{6.5}{8}\selectfont
\begingroup
\setlength{\tabcolsep}{0.45\tabcolsep}
\begin{tabular}{|c||c|c|c|c||c|c|c|c|}
\hline
\textbf{Dataset}  & \textbf{EEdesc}     & \textbf{EEasc}& \textbf{EEmix}& \textbf{EEunif}     & \textbf{EEdesc*}    & \textbf{EEasc*}     & \textbf{EEmix*}     & \textbf{EEunif*}    \\ \hline \hline
\textbf{Cifar10} & \textcolor{tabG}{3,795}    & \textcolor{tabG}{3,758}    & \textcolor{tabG}{5,232}    & \textcolor{tabG}{4,182}    & \textcolor{tabG}{4,261}    & \textcolor{tabG}{3,822}    & \textcolor{tabG}{\textbf{5,299}}    & \textcolor{tabG}{4,365}    \\ \hline
\textbf{Cifar100}& \textcolor{tabG}{13,972}   & \textcolor{tabG}{11,019}   & \textcolor{tabG}{16,117}   & \textcolor{tabG}{12,339}   & \textcolor{tabG}{15,888}   & \textcolor{tabG}{11,020}   & \textcolor{tabG}{\textbf{16,229}}   & \textcolor{tabG}{12,680}   \\ \hline
\textbf{Eurosat} & \textcolor{tabG}{3,027}    & \textcolor{tabG}{2,586}    & \textcolor{tabG}{2,996}    & \textcolor{tabG}{2,956}    & \textcolor{tabG}{3,046}    & \textcolor{tabG}{2,622}    & \textcolor{tabG}{\textbf{3,103}}    & \textcolor{tabG}{2,993}    \\ \hline
\textbf{FMNIST}  & \textcolor{tabG}{0,628}    & \textcolor{tabG}{0,912}    & \textcolor{tabG}{0,942}    & \textcolor{tabG}{1,122}    & \textcolor{tabG}{0,995}    & \textcolor{tabG}{0,926}    & \textcolor{tabG}{0,946}    & \textcolor{tabG}{\textbf{1,208}}    \\ \hline
\textbf{GTSRB}   & \textcolor{tabR}{-5,024}   & \textcolor{tabG}{1,719}    & \textcolor{tabR}{-2,914}   & \textcolor{tabR}{-0,746}   & \textcolor{tabR}{-3,846}   & \textcolor{tabG}{\textbf{1,873}}    & \textcolor{tabR}{-3,149}   & \textcolor{tabR}{-0,013}   \\ \hline
\textbf{TinyImg} & \textcolor{tabG}{11,462}   & \textcolor{tabG}{11,573}   & \textcolor{tabG}{13,586}   & \textcolor{tabG}{13,232}   & \textcolor{tabG}{\textbf{13,942}}   & \textcolor{tabG}{11,346}   & \textcolor{tabG}{13,867}   & \textcolor{tabG}{13,442}   \\ \hline \hline
\textbf{Total}   & \textcolor{tabG}{4,643} & \textcolor{tabG}{5,261} & \textcolor{tabG}{5,993} & \textcolor{tabG}{5,514} & \textcolor{tabG}{5,714} & \textcolor{tabG}{5,268} & \textcolor{tabG}{\textbf{6,049}} & \textcolor{tabG}{5,779} \\ \hline
\end{tabular}
\endgroup
    
\end{minipage}%
\vspace{0.4cm}
\hrule
\begin{center}
\textbf{TRAIN-64}
\end{center}
\hrule
\vspace{0.4cm}
\begin{minipage}{0.5\linewidth}

\fontsize{6.5}{8}\selectfont
\begingroup
\setlength{\tabcolsep}{0.45\tabcolsep}
\begin{tabular}{|c||c|c|c|c||c|c|c|c|}
\hline
\textbf{Network} & \textbf{EEdesc}& \textbf{EEasc}& \textbf{EEmix} & \textbf{EEunif}& \textbf{EEdesc*}     & \textbf{EEasc*}     & \textbf{EEmix*}& \textbf{EEunif*}     \\ \hline \hline
\textbf{Resnet50}    & \textcolor{tabG}{20,039}    & \textcolor{tabG}{10,444}   & \textcolor{tabG}{16,952}    & \textcolor{tabG}{14,368}    & \textcolor{tabG}{\textbf{20,642}}    & \textcolor{tabG}{10,655}   & \textcolor{tabG}{17,790}    & \textcolor{tabG}{14,402}    \\ \hline
\textbf{Vgg16} & \textcolor{tabG}{6,395}     & \textcolor{tabG}{4,273}    & \textcolor{tabG}{5,078}     & \textcolor{tabG}{5,704}     & \textcolor{tabG}{\textbf{6,942}}     & \textcolor{tabG}{4,252}    & \textcolor{tabG}{5,376}     & \textcolor{tabG}{5,607}     \\ \hline
\textbf{Vgg16full}   & \textcolor{tabG}{5,288}     & \textcolor{tabG}{4,359}    & \textcolor{tabG}{5,025}     & \textcolor{tabG}{5,743}     & \textcolor{tabG}{\textbf{6,277}}     & \textcolor{tabG}{4,452}    & \textcolor{tabG}{5,174}     & \textcolor{tabG}{5,909}     \\ \hline
\textbf{DenseNet169} & \textcolor{tabG}{9,852}     & \textcolor{tabG}{6,805}    & \textcolor{tabG}{8,222}     & \textcolor{tabG}{9,117}     & \textcolor{tabG}{\textbf{10,268}}    & \textcolor{tabG}{6,806}    & \textcolor{tabG}{8,751}     & \textcolor{tabG}{9,221}     \\ \hline
\textbf{MBV3-small} & \textcolor{tabG}{26,040}    & \textcolor{tabG}{16,323}   & \textcolor{tabG}{23,086}    & \textcolor{tabG}{20,008}    & \textcolor{tabG}{\textbf{26,129}}    & \textcolor{tabG}{16,377}   & \textcolor{tabG}{23,848}    & \textcolor{tabG}{19,956}    \\ \hline
\textbf{MBV3-smallfull}   & \textcolor{tabG}{26,940}    & \textcolor{tabG}{17,934}   & \textcolor{tabG}{24,223}    & \textcolor{tabG}{21,844}    & \textcolor{tabG}{\textbf{28,230}}    & \textcolor{tabG}{17,750}   & \textcolor{tabG}{25,559}    & \textcolor{tabG}{21,572}    \\ \hline
\textbf{EffNetB5}    & \textcolor{tabG}{7,367}     & \textcolor{tabG}{2,250}    & \textcolor{tabG}{6,616}     & \textcolor{tabG}{4,983}     & \textcolor{tabG}{7,477}     & \textcolor{tabG}{2,378}    & \textcolor{tabG}{\textbf{7,482}}     & \textcolor{tabG}{5,861}     \\ \hline
\textbf{EffNetB5full}& \textcolor{tabG}{13,607}    & \textcolor{tabG}{4,323}    & \textcolor{tabG}{11,213}    & \textcolor{tabG}{9,007}     & \textcolor{tabG}{\textbf{14,026}}    & \textcolor{tabG}{5,142}    & \textcolor{tabG}{12,500}    & \textcolor{tabG}{9,879}     \\ \hline \hline
\textbf{Total} & \textcolor{tabG}{14,441} & \textcolor{tabG}{8,339} & \textcolor{tabG}{12,552} & \textcolor{tabG}{11,347} & \textcolor{tabG}{\textbf{14,999}} & \textcolor{tabG}{8,477} & \textcolor{tabG}{13,310} & \textcolor{tabG}{11,551} \\ \hline
\end{tabular}
\endgroup
    
\end{minipage}%
\hspace{0.35cm}
\begin{minipage}{0.5\linewidth}

\fontsize{6.5}{8}\selectfont
\begingroup
\setlength{\tabcolsep}{0.45\tabcolsep}
\begin{tabular}{|c||c|c|c|c||c|c|c|c|}
\hline
\textbf{Dataset} & \textbf{EEdesc}& \textbf{EEasc}& \textbf{EEmix} & \textbf{EEunif}& \textbf{EEdesc*}     & \textbf{EEasc*}     & \textbf{EEmix*}& \textbf{EEunif*}     \\ \hline \hline
\textbf{Cifar10}     & \textcolor{tabG}{10,510}    & \textcolor{tabG}{5,643}    & \textcolor{tabG}{9,763}     & \textcolor{tabG}{8,233}     & \textcolor{tabG}{\textbf{10,595}}    & \textcolor{tabG}{5,811}    & \textcolor{tabG}{10,449}    & \textcolor{tabG}{8,333}     \\ \hline
\textbf{Cifat100}    & \textcolor{tabG}{24,093}    & \textcolor{tabG}{12,637}   & \textcolor{tabG}{21,284}    & \textcolor{tabG}{18,241}    & \textcolor{tabG}{\textbf{25,590}}    & \textcolor{tabG}{12,778}   & \textcolor{tabG}{22,409}    & \textcolor{tabG}{18,772}    \\ \hline
\textbf{Eurosat}     & \textcolor{tabG}{9,491}     & \textcolor{tabG}{4,462}    & \textcolor{tabG}{8,317}     & \textcolor{tabG}{7,316}     & \textcolor{tabG}{\textbf{9,774}}     & \textcolor{tabG}{5,051}    & \textcolor{tabG}{9,189}     & \textcolor{tabG}{7,718}     \\ \hline
\textbf{FMNIST}& \textcolor{tabG}{1,793}     & \textcolor{tabG}{1,071}    & \textcolor{tabG}{1,727}     & \textcolor{tabG}{1,772}     & \textcolor{tabG}{\textbf{1,804}}     & \textcolor{tabG}{1,062}    & \textcolor{tabG}{1,736}     & \textcolor{tabG}{1,734}     \\ \hline
\textbf{GTSRB} & \textcolor{tabG}{5,309}     & \textcolor{tabG}{7,208}    & \textcolor{tabG}{5,881}     & \textcolor{tabG}{6,853}     & \textcolor{tabG}{5,196}     & \textcolor{tabG}{\textbf{7,236}}    & \textcolor{tabG}{5,961}     & \textcolor{tabG}{6,860}     \\ \hline
\textbf{TinyImg}     & \textcolor{tabG}{35,450}    & \textcolor{tabG}{19,012}   & \textcolor{tabG}{28,340}    & \textcolor{tabG}{25,665}    & \textcolor{tabG}{\textbf{37,035}}    & \textcolor{tabG}{18,921}   & \textcolor{tabG}{30,115}    & \textcolor{tabG}{25,888}    \\ \hline \hline
\textbf{Total} & \textcolor{tabG}{14,441} & \textcolor{tabG}{8,339} & \textcolor{tabG}{12,552} & \textcolor{tabG}{11,347} & \textcolor{tabG}{\textbf{14,999}} & \textcolor{tabG}{8,477} & \textcolor{tabG}{13,310} & \textcolor{tabG}{11,551} \\ \hline
\end{tabular}
\endgroup

\end{minipage}%
\vspace{0.4cm}
\hrule
\begin{center}
\textbf{FINETUNE-224}
\end{center}
\hrule
\vspace{0.4cm}
\begin{minipage}{0.5\linewidth}

\fontsize{6.5}{8}\selectfont
\begingroup
\setlength{\tabcolsep}{0.45\tabcolsep}
\begin{tabular}{|c||c|c|c|c||c|c|c|c|}
\hline
 \textbf{Network}     & \textbf{EEdesc}& \textbf{EEasc}& \textbf{EEmix} & \textbf{EEunif}& \textbf{EEdesc*}     & \textbf{EEasc*}     & \textbf{EEmix*}& \textbf{EEunif*}    \\ \hline \hline
\textbf{Resnet50}  & \textcolor{tabR}{-4,199}    & \textcolor{tabG}{0,288}    & \textcolor{tabR}{-2,126}    & \textcolor{tabR}{-0,889}    & \textcolor{tabR}{-3,150}    & \textcolor{tabG}{\textbf{0,352}}    & \textcolor{tabR}{-2,218}    & \textcolor{tabR}{-0,490}   \\ \hline
\textbf{Vgg16}     & \textcolor{tabR}{-0,417}    & \textcolor{tabG}{1,756}    & \textcolor{tabG}{0,945}     & \textcolor{tabG}{1,845}     & \textcolor{tabG}{0,669}     & \textcolor{tabG}{1,709}    & \textcolor{tabG}{0,943}     & \textcolor{tabG}{\textbf{1,886}}    \\ \hline
\textbf{Vgg16full} & \textcolor{tabR}{-1,635}    & \textcolor{tabG}{1,895}    & \textcolor{tabG}{0,987}     & \textcolor{tabG}{1,550}     & \textcolor{tabG}{0,255}     & \textcolor{tabG}{1,977}    & \textcolor{tabG}{0,932}     & \textcolor{tabG}{\textbf{2,114}}    \\ \hline
\textbf{DenseNet169}     & \textcolor{tabR}{-6,263}    & \textcolor{tabR}{-1,196}   & \textcolor{tabR}{-4,153}    & \textcolor{tabR}{-2,520}    & \textcolor{tabR}{-4,870}    & \textcolor{tabR}{\textbf{-1,004}}   & \textcolor{tabR}{-3,961}    & \textcolor{tabR}{-2,018}   \\ \hline
\textbf{MBV3-small}     & \textcolor{tabR}{-3,813}    & \textcolor{tabG}{0,810}    & \textcolor{tabR}{-1,274}    & \textcolor{tabG}{0,119}     & \textcolor{tabR}{-1,691}    & \textcolor{tabG}{\textbf{0,833}}    & \textcolor{tabR}{-1,255}    & \textcolor{tabG}{0,593}    \\ \hline
\textbf{MBV3-smallfull} & \textcolor{tabR}{-4,619}    & \textcolor{tabG}{0,536}    & \textcolor{tabR}{-1,274}    & \textcolor{tabR}{-0,087}    & \textcolor{tabR}{-0,993}    & \textcolor{tabG}{\textbf{0,784}}    & \textcolor{tabR}{-1,164}    & \textcolor{tabG}{0,542}    \\ \hline
\textbf{EffNetB5}  & \textcolor{tabR}{-4,883}    & \textcolor{tabR}{-0,343}   & \textcolor{tabR}{-1,431}    & \textcolor{tabR}{-1,186}    & \textcolor{tabR}{-1,704}    & \textcolor{tabR}{\textbf{-0,228}}   & \textcolor{tabR}{-1,107}    & \textcolor{tabR}{-0,283}   \\ \hline
\textbf{EffNetB5full}    & \textcolor{tabR}{-4,920}    & \textcolor{tabR}{-0,604}   & \textcolor{tabR}{-1,956}    & \textcolor{tabR}{-1,342}    & \textcolor{tabR}{-2,535}    & \textcolor{tabR}{\textbf{-0,332}}   & \textcolor{tabR}{-1,809}    & \textcolor{tabR}{-0,649}   \\ \hline \hline
\textbf{Total}   & \textcolor{tabR}{-3,844} & \textcolor{tabG}{0,393} & \textcolor{tabR}{-1,285} & \textcolor{tabR}{-0,314} & \textcolor{tabR}{-1,752} & \textcolor{tabG}{\textbf{0,511}} & \textcolor{tabR}{-1,205} & \textcolor{tabG}{0,212} \\ \hline
\end{tabular}
\endgroup

\end{minipage}%
\hspace{0.35cm}
\begin{minipage}{0.5\linewidth}

\fontsize{6.5}{8}\selectfont
\begingroup
\setlength{\tabcolsep}{0.45\tabcolsep}
\begin{tabular}{|c||c|c|c|c||c|c|c|c|}
\hline
\textbf{Dataset}& \textbf{EEdesc}& \textbf{EEasc}& \textbf{EEmix} & \textbf{EEunif}& \textbf{EEdesc*}     & \textbf{EEasc*}     & \textbf{EEmix*}& \textbf{EEunif*}    \\ \hline \hline
\textbf{Cifar10} & \textcolor{tabR}{-3,247}    & \textcolor{tabG}{0,070}    & \textcolor{tabR}{-1,000}    & \textcolor{tabR}{-0,690}    & \textcolor{tabR}{-1,393}    & \textcolor{tabG}{\textbf{0,260}}    & \textcolor{tabR}{-0,892}    & \textcolor{tabR}{-0,229}   \\ \hline
\textbf{Cifar100}& \textcolor{tabR}{-6,428}    & \textcolor{tabG}{0,973}    & \textcolor{tabR}{-1,718}    & \textcolor{tabG}{0,268}     & \textcolor{tabR}{-2,540}    & \textcolor{tabG}{\textbf{1,015}}    & \textcolor{tabR}{-1,593}    & \textcolor{tabG}{0,867}    \\ \hline
\textbf{Eurosat} & \textcolor{tabG}{0,492}     & \textcolor{tabG}{\textbf{0,812}}    & \textcolor{tabG}{0,556}     & \textcolor{tabG}{0,698}     & \textcolor{tabG}{0,627}     & \textcolor{tabG}{0,805}    & \textcolor{tabG}{0,556}     & \textcolor{tabG}{0,701}    \\ \hline
\textbf{FMNIST}  & \textcolor{tabR}{-0,485}    & \textcolor{tabG}{\textbf{0,040}}    & \textcolor{tabR}{-0,065}    & \textcolor{tabR}{-0,030}    & \textcolor{tabR}{-0,096}    & \textcolor{tabG}{0,039}    & \textcolor{tabR}{-0,090}    & \textcolor{tabG}{0,030}    \\ \hline
\textbf{GTSRB}   & \textcolor{tabR}{-2,653}    & \textcolor{tabR}{-0,207}   & \textcolor{tabR}{-0,940}    & \textcolor{tabR}{-0,787}    & \textcolor{tabR}{-1,529}    & \textcolor{tabR}{\textbf{-0,201}}   & \textcolor{tabR}{-0,919}    & \textcolor{tabR}{-0,354}   \\ \hline
\textbf{TinyImg} & \textcolor{tabR}{-10,742}   & \textcolor{tabG}{0,669}    & \textcolor{tabR}{-4,546}    & \textcolor{tabR}{-1,340}    & \textcolor{tabR}{-5,583}    & \textcolor{tabG}{\textbf{1,150}}    & \textcolor{tabR}{-4,292}    & \textcolor{tabG}{0,257}    \\ \hline \hline
\textbf{Total}   & \textcolor{tabR}{-3,844} & \textcolor{tabG}{0,393} & \textcolor{tabR}{-1,285} & \textcolor{tabR}{-0,314} & \textcolor{tabR}{-1,752} & \textcolor{tabG}{\textbf{0,511}} & \textcolor{tabR}{-1,205} & \textcolor{tabG}{0,212} \\ \hline
\end{tabular}
\endgroup

\end{minipage}%
\vspace{0.4cm}
\hrule
\begin{center}
\textbf{FINETUNE-64}
\end{center}
\hrule
\vspace{0.4cm}
\begin{minipage}{0.5\linewidth}

\fontsize{6.5}{8}\selectfont
\begingroup
\setlength{\tabcolsep}{0.45\tabcolsep}
\begin{tabular}{|c||c|c|c|c||c|c|c|c|}
\hline
\textbf{Network}  & \textbf{EEdesc}     & \textbf{EEasc}& \textbf{EEmix}& \textbf{EEunif}     & \textbf{EEdesc*}    & \textbf{EEasc*}     & \textbf{EEmix*}     & \textbf{EEunif*}    \\ \hline \hline
\textbf{Resnet50}  & \textcolor{tabG}{1,048}    & \textcolor{tabG}{\textbf{4,252}}    & \textcolor{tabG}{0,861}    & \textcolor{tabG}{3,687}    & \textcolor{tabG}{1,233}    & \textcolor{tabG}{3,895}    & \textcolor{tabG}{0,697}    & \textcolor{tabG}{3,660}    \\ \hline
\textbf{Vgg16}     & \textcolor{tabG}{2,516}    & \textcolor{tabG}{\textbf{3,500}}    & \textcolor{tabG}{2,148}    & \textcolor{tabG}{2,555}    & \textcolor{tabG}{2,811}    & \textcolor{tabG}{3,403}    & \textcolor{tabG}{2,151}    & \textcolor{tabG}{2,584}    \\ \hline
\textbf{Vgg16full} & \textcolor{tabG}{1,805}    & \textcolor{tabG}{3,714}    & \textcolor{tabG}{2,108}    & \textcolor{tabG}{3,454}    & \textcolor{tabG}{2,738}    & \textcolor{tabG}{\textbf{3,762}}    & \textcolor{tabG}{2,110}    & \textcolor{tabG}{3,508}    \\ \hline
\textbf{DenseNet169}     & \textcolor{tabR}{-0,399}   & \textcolor{tabG}{2,790}    & \textcolor{tabG}{0,113}    & \textcolor{tabG}{1,985}    & \textcolor{tabR}{-0,156}   & \textcolor{tabG}{\textbf{2,979}}    & \textcolor{tabG}{0,190}    & \textcolor{tabG}{2,211}    \\ \hline
\textbf{MBV3-small}     & \textcolor{tabG}{6,060}    & \textcolor{tabG}{6,084}    & \textcolor{tabG}{5,949}    & \textcolor{tabG}{\textbf{6,646}}    & \textcolor{tabG}{6,417}    & \textcolor{tabG}{5,850}    & \textcolor{tabG}{6,009}    & \textcolor{tabG}{6,585}    \\ \hline
\textbf{MBV3-smallfull} & \textcolor{tabG}{5,097}    & \textcolor{tabG}{6,119}    & \textcolor{tabG}{5,773}    & \textcolor{tabG}{6,233}    & \textcolor{tabG}{6,418}    & \textcolor{tabG}{6,102}    & \textcolor{tabG}{5,719}    & \textcolor{tabG}{\textbf{6,500}}    \\ \hline
\textbf{EffNetB5}  & \textcolor{tabG}{5,181}    & \textcolor{tabG}{5,768}    & \textcolor{tabG}{5,137}    & \textcolor{tabG}{6,040}    & \textcolor{tabG}{5,243}    & \textcolor{tabG}{5,771}    & \textcolor{tabG}{5,126}    & \textcolor{tabG}{\textbf{6,045}}    \\ \hline
\textbf{EffNetB5full}    & \textcolor{tabG}{6,189}    & \textcolor{tabG}{6,075}    & \textcolor{tabG}{6,120}    & \textcolor{tabG}{6,771}    & \textcolor{tabG}{6,291}    & \textcolor{tabG}{6,123}    & \textcolor{tabG}{6,397}    & \textcolor{tabG}{\textbf{6,792}}    \\ \hline \hline
\textbf{Total}     & \textcolor{tabG}{3,437} & \textcolor{tabG}{\textbf{4,788}} & \textcolor{tabG}{3,526} & \textcolor{tabG}{4,671} & \textcolor{tabG}{3,874} & \textcolor{tabG}{4,736} & \textcolor{tabG}{3,550} & \textcolor{tabG}{4,736} \\ \hline
\end{tabular}
\endgroup

\end{minipage}%
\hspace{0.35cm}
\begin{minipage}{0.5\linewidth}

\fontsize{6.5}{8}\selectfont
\begingroup
\setlength{\tabcolsep}{0.45\tabcolsep}
\begin{tabular}{|c||c|c|c|c||c|c|c|c|}
\hline
\textbf{Dataset}  & \textbf{EEdesc}     & \textbf{EEasc}& \textbf{EEmix}& \textbf{EEunif}     & \textbf{EEdesc*}    & \textbf{EEasc*}     & \textbf{EEmix*}     & \textbf{EEunif*}    \\ \hline \hline
\textbf{Cifar10} & \textcolor{tabG}{2,722}    & \textcolor{tabG}{3,401}    & \textcolor{tabG}{2,983}    & \textcolor{tabG}{3,737}    & \textcolor{tabG}{2,966}    & \textcolor{tabG}{3,415}    & \textcolor{tabG}{3,063}    & \textcolor{tabG}{\textbf{3,768}}    \\ \hline
\textbf{Cifar100}& \textcolor{tabG}{5,402}    & \textcolor{tabG}{7,594}    & \textcolor{tabG}{5,835}    & \textcolor{tabG}{7,876}    & \textcolor{tabG}{6,123}    & \textcolor{tabG}{7,563}    & \textcolor{tabG}{5,891}    & \textcolor{tabG}{\textbf{7,954}}    \\ \hline
\textbf{Eurosat} & \textcolor{tabG}{3,480}    & \textcolor{tabG}{3,186}    & \textcolor{tabG}{3,298}    & \textcolor{tabG}{3,395}    & \textcolor{tabG}{\textbf{3,512}}    & \textcolor{tabG}{3,230}    & \textcolor{tabG}{3,289}    & \textcolor{tabG}{3,494}    \\ \hline
\textbf{FMNIST}  & \textcolor{tabG}{0,835}    & \textcolor{tabG}{0,916}    & \textcolor{tabG}{1,043}    & \textcolor{tabG}{1,093}    & \textcolor{tabG}{0,926}    & \textcolor{tabG}{0,907}    & \textcolor{tabG}{1,093}    & \textcolor{tabG}{\textbf{1,138}}    \\ \hline
\textbf{GTSRB}   & \textcolor{tabG}{4,433}    & \textcolor{tabG}{\textbf{5,640}}    & \textcolor{tabG}{4,571}    & \textcolor{tabG}{5,136}    & \textcolor{tabG}{4,712}    & \textcolor{tabG}{5,355}    & \textcolor{tabG}{4,417}    & \textcolor{tabG}{5,070}    \\ \hline
\textbf{TinyImg} & \textcolor{tabG}{3,750}    & \textcolor{tabG}{\textbf{7,989}}    & \textcolor{tabG}{3,426}    & \textcolor{tabG}{6,791}    & \textcolor{tabG}{5,007}    & \textcolor{tabG}{7,943}    & \textcolor{tabG}{3,547}    & \textcolor{tabG}{6,990}    \\ \hline \hline
\textbf{Total}     & \textcolor{tabG}{3,437} & \textcolor{tabG}{\textbf{4,788}} & \textcolor{tabG}{3,526} & \textcolor{tabG}{4,671} & \textcolor{tabG}{3,874} & \textcolor{tabG}{4,736} & \textcolor{tabG}{3,550} & \textcolor{tabG}{4,736} \\ \hline
\end{tabular}
\endgroup

\end{minipage}%

\label{table:grouped_top1}
\end{table*}

This section describes the results of the experiments conducted in the present research. The analysis begins by comparing the variations in accuracy between single-output architectures and the proposed configurations. Next, the benefits of the pruning step are shown, in terms of accuracy gain, parameter reduction, optimisation of MACs and faster inference time. Finally, the accuracy performance of early exits without ensemble compared to single-output models is analysed. The symbol '*' indicates early exits experiments whose networks have been subjected to pruning. In contrast, the absence of the symbol '*' represents full-ensemble networks.

\subsection{Classification Accuracy}

Table~\ref{table:grouped_top1} shows the results of the experiments performed in terms of the percentage change in Top1 accuracy compared to the single-output experiments, grouped by neural network in the left column and by dataset in the right column.

With regard to the experiments performed with 224x224 images and traditional training (TRAIN-224), excellent average improvements can first of all be appreciated regardless of the strategy with which AEP was applied. For each neural network and each dataset, it is possible to identify a configuration that improves the accuracy obtained. In particular, as far as the datasets are concerned, the greatest advantages are found in CIFAR100 and TinyImageNet, which respectively obtain an average accuracy improvement of 16.23\% with the EEmix* configuration and 13.94\% with the EEdesc* configuration. With regard to network configurations, the improvement achieved by ResNet50 and both MobileNetV3-small versions is noteworthy.

The general considerations made for the TRAIN-224 scenario are confirmed and underlined by the results of the experiments with 64x64 images and traditional training (TRAIN-64). In this case, with the same dataset and neural network, the problem to be solved by the algorithm considered is more complex due to the reduced amount of input information. In this context, it is possible to observe how beneficial AEP is overall, increasing the quality of the models regardless of their configuration and dataset, in some cases making such improvements as to make the difference between a poorly performing model and a winning one. 
In this scenario, the EEdesc* technique seems to be definitely the favourite, capable of improving accuracy performance by 15\% on average.

Moving from the training scenario to the fine-tuning scenario, and in particular to the one with 224x224 images (FINETUNE-224), a different behavior can be observed. Since these are pre-trained algorithms on ImageNet images with similar size, the addition of early exits and ensembles seems to have a negative overall impact on average. This is reasonable in light of the fact that the starting models have an optimized set of weights available to facilitate the production of accurate predictions from the single output at the bottom of the network. This assumption is confirmed by the counterexample represented by the results obtained from the EEasc and EEasc* configurations. In fact, it can be observed from Table~\ref{table:grouped_top1} how the use of ascending weights in the loss and inference phase turns out to be more suitable by favoring strongly the contribution of the outputs close to the single output of the original model.

Finally, going to look at the results of the last pool of experiments,i.e., those with fine-tuning and 64x64 images (FINETUNE-64), it is possible to see a different behavior from the previous case. The change of spatial domain resulted in an overall improvement in the accuracy of the AEP-based models compared to the single-output counterpart. This is probably due to the higher complexity of the problem inversely proportional to the size of the input, which thus allows better exploitation of classifiers trained on lower-level and thus less elaborate features, as is not the case with large images that require more complex transformations. However, the use of ascending weights turns out to be the winning move here as well, capable of improving models accuracy by an average of 4.79\%.

\begin{table}[t]
\caption{Comparisons between models based on early exits after pruning and their counterparts with full-ensemble, calculated as average percentage change of accuracy. The four experiment configurations are divided by row according to training technique and input image size. Improvements are shown in green, deteriorations in red. The best results are highlighted in bold. The obtained values are averaged over architectures and datasets.
}
\begin{center}
\begin{tabular}{|p{1.3cm}||p{1.3cm}|p{1.3cm}|p{1.3cm}|p{1.3cm}|}
\hline
  & \textbf{EEdesc*} & \textbf{EEasc*}   & \textbf{EEmix*}  & \textbf{EEunif*}\\ \hline \hline
\textbf{TR-224}   & \textcolor{tabG}{\textbf{1,006}} & \textcolor{tabG}{0,009}  & \textcolor{tabG}{0,050} & \textcolor{tabG}{0,260} \\ 
\textbf{TR-64}    & \textcolor{tabG}{0,435} & \textcolor{tabG}{0,142}  & \textcolor{tabG}{\textbf{0,625}} & \textcolor{tabG}{0,193} \\ 
\textbf{FT-224}   & \textcolor{tabG}{\textbf{2,280}} & \textcolor{tabG}{0,121}  & \textcolor{tabG}{0,087} & \textcolor{tabG}{0,539} \\ 
\textbf{FT-64}    & \textcolor{tabG}{\textbf{0,421}} & \textcolor{tabR}{-0,048} & \textcolor{tabG}{0,021} & \textcolor{tabG}{0,063} \\ \hline
\end{tabular}
\label{table:me_vs_mestar_top1}
\end{center}
\end{table}

In general, it can be argued that ensemble networks with early exits improve much more than the baseline when trained from scratch than when fine-tuned. 
This is reasonable because when training from scratch by jointly optimising the exits, the network is able to improve the features of each layer to achieve a good classification result, producing better features in the layers closer to the network input. 
Fine-tuning, on the other hand, starts from the weights obtained after training single-output networks, which means that the features of the initial layers are not good classification features by themselves, but do produce good classification features in the final layers. 
Adding early exits therefore leads to contradictory behaviour, as if the optimisation were trying to override the initial weights to solve a different task.
In fact, by assigning higher weights to exits closer to the input, you end up assigning high classification importance to layers that should not be able to do so, which can lead to worse performance than the baseline.

Regarding the effect of the pruning step, the same behaviour as described in full ensembles can be observed with slight improvements in accuracy.
Table~\ref{table:me_vs_mestar_top1} collects and compares the accuracy performance of pruned versions of early exits ensemble networks with their full-ensemble counterparts.
Among the different types of networks, EEdesc networks seem to be the ones that benefit the most from the pruning phase, particularly in fine-tuning, the same learning context in which they perform the worst.

\subsection{Parameters, Operations and Latency}

The average effect in terms of parameters of applying AEP is summarised in Table~\ref{table:me_vs_se_params}.
As can be seen from the first column, the addition of early exits has a minimal impact on the increase in parameters, which on average increases by only 2.88\% compared to single-exit architectures.
The key aspect is the impact of the pruning step, which in general brings considerable benefits. Paying attention to the TRAINING-64 scenario, it can be seen that the EEmix* configuration results in an average parameter reduction of 41.02\% and in parallel an average accuracy gain of 13.31\%, as described in Table~\ref{table:grouped_top1}.

With regard to the number of MACs, the average results of the experiments conducted are summarised in Table~\ref{table:me_vs_se_macs}. Also for this metric, the addition of early exits alone has a negative impact, albeit with an average computational increase of only 5\%. On the contrary, it is possible to observe the beneficial effect of pruning, which in addition to providing the previously discussed benefits, goes as far as reducing the number of operations by up to a maximum of 17.95\% in the EEmix* case.

\begin{table}[t]
\caption{Comparisons between models based on early exits and their single-exit counterparts, calculated as average percentage change of parameters number. The four experiment configurations are divided by row according to training technique and input image size. Improvements are shown in green, deteriorations in red. The best results are highlighted in bold. The obtained values are averaged over architectures and datasets.}
\begin{center}
\begin{tabular}{|p{1.2cm}||p{1cm}|p{1cm}|p{1cm}|p{1cm}|p{1cm}|}
\hline
    & \textbf{EE} & \textbf{EEdesc*}   & \textbf{EEasc*}    & \textbf{EEmix*}    & \textbf{EEunif*}               \\ \hline \hline
\textbf{TR-224} & \textcolor{tabR}{2,876}       & \textcolor{tabG}{-0,566}  & \textcolor{tabG}{-10,422} & \textcolor{tabG}{\textbf{-16,508}} & \textcolor{tabG}{-1,773}  \\ 
\textbf{TR-64}  & \textcolor{tabR}{2,876}       & \textcolor{tabG}{-12,731} & \textcolor{tabG}{-22,549} & \textcolor{tabG}{\textbf{-41,018}} & \textcolor{tabG}{-25,300} \\ 
\textbf{FT-224} & \textcolor{tabR}{2,876}       & \textcolor{tabG}{\textbf{-5,791}}  & \textcolor{tabR}{1,773}   & \textcolor{tabG}{-1,902}  & \textcolor{tabG}{-0,720}  \\ 
\textbf{FT-64}  & \textcolor{tabR}{2,876}       & \textcolor{tabG}{-4,260}  & \textcolor{tabG}{-14,305} & \textcolor{tabG}{\textbf{-19,351}} & \textcolor{tabG}{-11,292} \\ \hline

\end{tabular}
\label{table:me_vs_se_params}
\end{center}
\end{table}
\begin{table}[t]
\caption{Comparisons between models based on early exits and their single-exit counterparts, calculated as average percentage change of MACs. The four experiment configurations are divided by row according to training technique and input image size. Improvements are shown in green, deteriorations in red. The best results are highlighted in bold. The obtained values are averaged over architectures and datasets.}
\begin{center}
\begin{tabular}{|p{1cm}||p{1cm}|p{1cm}|p{1cm}|p{1cm}|p{1cm}|}
\hline
    & \textbf{EE} & \textbf{EEdesc*}  & \textbf{EEasc*}   & \textbf{EEmix*}    & \textbf{EEunif*}              \\ \hline \hline
\textbf{TR-224}   & \textcolor{tabR}{5,444}       & \textcolor{tabR}{1,816}  & \textcolor{tabG}{-1,908} & \textcolor{tabG}{\textbf{-5,438}}  & \textcolor{tabR}{2,118}  \\ 
\textbf{TR-64}    & \textcolor{tabR}{5,260}       & \textcolor{tabG}{-3,302} & \textcolor{tabG}{-8,549} & \textcolor{tabG}{\textbf{-17,952}} & \textcolor{tabG}{-9,078} \\ 
\textbf{FT-224}   & \textcolor{tabR}{5,444}       & \textcolor{tabG}{\textbf{-1,475}} & \textcolor{tabR}{2,830}  & \textcolor{tabR}{2,514}   & \textcolor{tabR}{1,115}  \\ 
\textbf{FT-64}    & \textcolor{tabR}{5,260}       & \textcolor{tabR}{0,059}  & \textcolor{tabG}{-1,954} & \textcolor{tabG}{\textbf{-4,085}}  & \textcolor{tabG}{-1,050} \\ \hline

\end{tabular}
\label{table:me_vs_se_macs}
\end{center}
\end{table}
\begin{table}[t]
\caption{Comparisons between models based on early exits and their single-exit counterparts, calculated as average percentage change of inference latency. The four experiment configurations are divided by row according to training technique and input image size. Improvements are shown in green, deteriorations in red. The best results are highlighted in bold. The obtained values are averaged over architectures and datasets.}
\begin{center}
\begin{tabular}{|p{1cm}||p{1cm}|p{1cm}|p{1cm}|p{1cm}|p{1cm}|}
\hline
    & \textbf{EE} & \textbf{EEdesc*} & \textbf{EEasc*}  & \textbf{EEmix*}  & \textbf{EEunif*}    \\ \hline \hline
\textbf{TR-224} & \textcolor{tabR}{5,795}  & \textcolor{tabR}{3,782} & \textcolor{tabR}{0,387}  & \textcolor{tabG}{\textbf{-2,729}}  & \textcolor{tabR}{3,220}  \\
\textbf{TR-64}  & \textcolor{tabR}{7,771}  & \textcolor{tabG}{-0,776}& \textcolor{tabG}{-6,061} & \textcolor{tabG}{\textbf{-16,325}} & \textcolor{tabG}{-7,482} \\
\textbf{FT-224} & \textcolor{tabR}{6,645}  & \textcolor{tabR}{\textbf{0,407}} & \textcolor{tabR}{4,959}  & \textcolor{tabR}{4,177}   & \textcolor{tabR}{3,143}  \\
\textbf{FT-64}  & \textcolor{tabR}{7,877}  & \textcolor{tabR}{2,264} & \textcolor{tabG}{-1,511} & \textcolor{tabG}{\textbf{-3,111}}  & \textcolor{tabR}{0,122}  \\ \hline

\end{tabular}
\label{table:me_vs_se_latency}
\end{center}
\end{table}

A similar behaviour can be observed in the inference latency domain, as shown by the results of the experiments summarised in Table~\ref{table:me_vs_se_latency}.
Adding more outputs and calculating their weighted sum is clearly disadvantageous from a time point of view. In percentage terms, this phenomenon is more pronounced in architectures with input sizes 64x64. The cut of unnecessary outputs implemented by the pruning operation inevitably benefits inference times, as the image input to the networks will on average have to travel through a portion of the architecture and not its entirety. Fine-tuning scenarios tend to benefit less from the pruning step precisely because, as can be deduced from Table~\ref{table:me_vs_se_params} and Table~\ref{table:me_vs_se_macs}, on average only the very last outputs, or none at all, are cut out. Consequently, although mitigated, the addition of early exits is only partially beneficial, as opposed to scenarios characterised by training from scratch.

\subsection{Early Exits Analysis}

An interesting aspect is certainly that of the quality of the classifiers obtained prior to the pruning operation. In fact, the proposed training, whether from scratch or fine-tuning, has the purpose of optimising the ensemble to be pruned, but has the secondary effect of computing weights that can be used directly by the classifiers obtained thanks to early exits.
For each of the scenarios considered, Table~\ref{table:se_vs_me_exits} presents the average percentage change in accuracy of the early outputs common to all configurations compared to the average performance achieved by single-output networks.

\begin{table}[t]
\caption{Comparisons in terms of percentage change of accuracy between the individual outputs of the models based on early exits and their single-output counterparts. The four experiment configurations are divided by table according to training technique and input image size. Improvements are shown in green, deteriorations in red. Best results are highlighted in bold. The obtained values are averaged over architectures and datasets.}
\vspace{5mm}
\centering

    \begin{tabular}{|p{1.3cm}||p{1.3cm}|p{1.3cm}|p{1.3cm}|p{1.3cm}|p{1.3cm}|}
    \hline
    \textbf{TR-224}  & \textbf{exit1}     & \textbf{exit2}    & \textbf{exit3}   & \textbf{exit4}               \\ \hline \hline
    \textbf{EEdesc} & \textcolor{tabR}{-30,155} & \textcolor{tabR}{-4,193}  & \textcolor{tabG}{2,508} & \textcolor{tabG}{\textbf{2,793}} \\
    \textbf{EEasc}  & \textcolor{tabR}{-42,465} & \textcolor{tabR}{-12,538} & \textcolor{tabG}{1,171} & \textcolor{tabG}{\textbf{3,041}} \\
    \textbf{EEmix}  & \textcolor{tabR}{-31,434} & \textcolor{tabR}{-5,444}  & \textcolor{tabG}{2,884} & \textcolor{tabG}{\textbf{3,115}} \\
    \textbf{EEunif} & \textcolor{tabR}{-35,224} & \textcolor{tabR}{-8,517}  & \textcolor{tabG}{1,942} & \textcolor{tabG}{\textbf{2,970}} \\ \hline
    \end{tabular}

\vspace{3mm}

    \begin{tabular}{|p{1.3cm}||p{1.3cm}|p{1.3cm}|p{1.3cm}|p{1.3cm}|p{1.3cm}|}
    \hline
    \textbf{TR-64}  & \textbf{exit1}     & \textbf{exit2}    & \textbf{exit3}   & \textbf{exit4}               \\ \hline \hline
    \textbf{EEdesc} & \textcolor{tabR}{-10,432} & \textcolor{tabG}{\textbf{9,515}}  & \textcolor{tabG}{9,442} & \textcolor{tabG}{7,739} \\
    \textbf{EEasc}  & \textcolor{tabR}{-25,613} & \textcolor{tabR}{-1,243} & \textcolor{tabG}{\textbf{4,999}} & \textcolor{tabG}{4,373} \\
    \textbf{EEmix}  & \textcolor{tabR}{-11,434} & \textcolor{tabG}{8,350}  & \textcolor{tabG}{\textbf{9,957}} & \textcolor{tabG}{8,060} \\
    \textbf{EEunif} & \textcolor{tabR}{-17,433} & \textcolor{tabG}{4,527}  & \textcolor{tabG}{\textbf{7,027}} & \textcolor{tabG}{5,692} \\ \hline
    \end{tabular}

\vspace{3mm}

    \begin{tabular}{|p{1.3cm}||p{1.3cm}|p{1.3cm}|p{1.3cm}|p{1.3cm}|p{1.3cm}|}
    \hline
    \textbf{FT-224} & \textbf{exit1}            & \textbf{exit2}            & \textbf{exit3}            & \textbf{exit4}                \\ \hline \hline
    \textbf{EEdesc} & \textcolor{tabR}{-37,867} & \textcolor{tabR}{-13,969} & \textcolor{tabR}{-5,713} & \textcolor{tabR}{\textbf{-3,027}} \\
    \textbf{EEasc}  & \textcolor{tabR}{-48,667} & \textcolor{tabR}{-19,457} & \textcolor{tabR}{-5,634} & \textcolor{tabR}{\textbf{-0,528}} \\
    \textbf{EEmix}  & \textcolor{tabR}{-39,664} & \textcolor{tabR}{-16,120} & \textcolor{tabR}{-5,777} & \textcolor{tabR}{\textbf{-2,574}} \\
    \textbf{EEunif} & \textcolor{tabR}{-41,946} & \textcolor{tabR}{-16,633} & \textcolor{tabR}{-5,089} & \textcolor{tabR}{\textbf{-1,066}} \\ \hline
    \end{tabular}

\vspace{3mm}

    \begin{tabular}{|p{1.3cm}||p{1.3cm}|p{1.3cm}|p{1.3cm}|p{1.3cm}|p{1.3cm}|}
    \hline
    \textbf{FT-64}  & \textbf{exit1}            & \textbf{exit2}    & \textbf{exit3}   & \textbf{exit4}               \\ \hline \hline
    \textbf{EEdesc} & \textcolor{tabR}{-20,932} & \textcolor{tabR}{-2,276} & \textcolor{tabG}{\textbf{1,181}} & \textcolor{tabG}{0,649} \\
    \textbf{EEasc}  & \textcolor{tabR}{-32,504} & \textcolor{tabR}{-7,237} & \textcolor{tabG}{1,902} & \textcolor{tabG}{\textbf{2,773}} \\
    \textbf{EEmix}  & \textcolor{tabR}{-22,601} & \textcolor{tabR}{-3,186} & \textcolor{tabG}{\textbf{1,173}} & \textcolor{tabG}{0,874} \\
    \textbf{EEunif} & \textcolor{tabR}{-25,106} & \textcolor{tabR}{-3,932} & \textcolor{tabG}{1,891} & \textcolor{tabG}{\textbf{1,915}} \\ \hline
    \end{tabular}

\label{table:se_vs_me_exits}
\end{table}

The first notable result is the presence of early exits performing better than the single output network. This occurs in every scenario except FINETUNE-224, managing to achieve average accuracy improvements of up to 9.96\%. This confirms not only the effectiveness of AEP as an ensemble-based training technique, but also opens up the possibility of realising shallower and at the same time more performant architectures, particularly useful in mobile or edge or distributed setups.
Focusing on the TRAIN-64 and TRAIN-224 scenarios, it is evident that both third and fourth outputs represent an advantageous alternative to the traditional single output regardless of the AEP configuration used. This result is perhaps the most significant. It is important to reflect on the fact that sometimes good human intuitions, such as increasing the complexity of a model, do not necessarily translate into improvements in performance. On the contrary, better results can be achieved by optimising the learning process and exploiting various levels of abstraction of the input data, as is done through the use of multiple outputs.

\section{Conclusion}\label{sec:conclusion}

This paper presented Anticipate, Ensemble and Prune (AEP), the early exits and ensemble-based technique for improving the performance of single-output artificial neural networks. The extensive series of experiments conducted demonstrated how this approach can be applied in the context of image classification, achieving remarkable improvements in terms of accuracy, parametric complexity, number of operations and inference time. The experiments were also replicated in the context of fitting pre-trained networks using fine-tuning techniques, with excellent results, especially in the absence of high-resolution input images.
It was also shown how AEP can optimise the performance of individual early exits, exceeding the accuracy of single-exit models even without using the ensemble technique.
We believe this can be an important step towards optimising the efficiency of neural architectures, especially for mobile and edge scenarios.
In a follow-up study, AEP will include an automatic optimisation process to find the best loss weights and a search process to find the best output weights, which is likely to lead to further improvements and potentially reveal undiscovered design patterns.

\section*{Acknowledgment}
The European Commission has partially funded this work under the H2020 grant N. 101016577 AI-SPRINT: AI in Secure Privacy-pReserving computINg conTinuum.

\bibliographystyle{IEEEtran}
\bibliography{references}
\end{document}